\pdfoutput=1
\documentclass{llncs}
\usepackage{llncsdoc}
\usepackage{amsmath}
\usepackage{graphicx}
\usepackage{amsfonts}
\usepackage{amsmath}
\usepackage{subfig}
\usepackage{soul}
\usepackage [english]{babel}
\usepackage [autostyle, english = american]{csquotes}
\MakeOuterQuote{"}
\usepackage{hyperref}

\graphicspath{ {Images/} }

\begin{document}

%% Author and institute
\title{Self-supervised Learning for Dense Depth Estimation in Monocular Endoscopy}

\author{Xingtong Liu \inst{1} \and Ayushi Sinha \inst{1} \and Mathias Unberath \inst{1} \and Masaru Ishii \inst{2} \and Gregory D. Hager \inst{1} \and Russell H. Taylor \inst{1} \thanks{Russell H. Taylor is a paid consultant to and owns equity in Galen Robotics, Inc. These arrangements have been reviewed and approved by JHU in accordance with its conflict of interest policy.} \and Austin Reiter \inst{1}}

\institute{The Johns Hopkins University, Baltimore, USA,\\ \email{xliu89@jh.edu}
\and
Johns Hopkins Medical Institutions, Baltimore, USA}

\maketitle

%% Abstract
\begin{abstract}
We present a self-supervised approach to training convolutional neural networks for dense depth estimation from monocular endoscopy data without \textit{a priori} modeling of anatomy or shading. Our method only requires sequential data from monocular endoscopic videos and a multi-view stereo reconstruction method, e.g. structure from motion, that supervises learning in a sparse but accurate manner. Consequently, our method requires neither manual interaction, such as scaling or labeling, nor patient CT in the training and application phases. We demonstrate the performance of our method on sinus endoscopy data from two patients and validate depth prediction quantitatively using corresponding patient CT scans where we found submillimeter residual errors.\footnote{Link to the supplementary video: \url{https://camp.lcsr.jhu.edu/miccai-2018-demonstration-videos/}}
\end{abstract}

\section{Introduction}
\begin{figure}[t]
	\centering
	\includegraphics[scale=0.2]{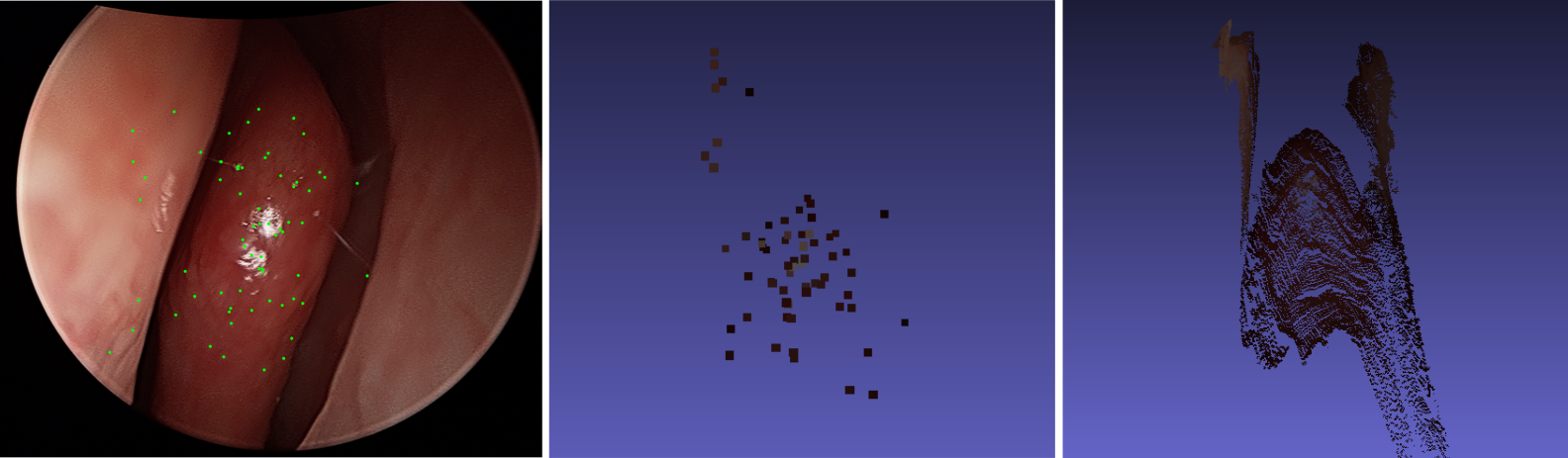}
	\caption{Visual comparison of reconstructions: the green dots in the endoscopic image (left) are 2D projections of the sparse reconstruction (middle) from a recent SfM-based method~\cite{Simon:SfM}. In this example, SfM only yields 67 3D points. Our method (right) produces a dense reconstruction with 125369 3D points, shown here from approximately the same viewpoint as the SfM reconstruction. The higher the resolution of the input image, the greater the number of points our method is able to reconstruct.}
	\label{fig:comparison}
\end{figure}

Minimally invasive procedures, such as functional endoscopic sinus surgery, typically employ surgical navigation systems to visualize critical structures that must not be disturbed during surgery. Computer vision-based navigation systems that rely on endoscopic video and do not introduce additional hardware are both easy to integrate into clinical workflow and cost effective. Such systems generally rely on the registration of preoperative data, such as CT scans, to intraoperative endoscopic video data~\cite{Simon:SfM}. This registration must be highly accurate in order to guarantee reliable performance of the navigation system. Since the accuracy of feature-based video-CT registration methods is dependent on the quality of reconstructions obtained from endoscopic video, it is critical for these reconstructions to be accurate. Further, in order to solve for the additional degrees of freedom required by deformable registration methods~\cite{Sinha18}, these reconstructions must also be dense. Our method satisfies both of these requirements (Fig.~\ref{fig:comparison}). %In other applications like surgical force prediction, dense reconstructions of anatomy can also be useful, as shown in~\cite{Cong18}.
%Most commonly, registration is approached in 3D by registering the CT data to interventional reconstructions of the anatomy from endoscopic video suggesting that reconstruction quality will substantially affect registration performance. 

Several reconstruction methods have been explored in the past. Multi-view stereo methods, such as Structure from Motion (SfM)~\cite{Simon:SfM} and Simultaneous Localization and Mapping (SLAM)~\cite{Grasa:SLAM}, are able to simultaneously reconstruct 3D structure and estimate camera poses in feature-rich scenes. However, the paucity of features in endoscopic images of anatomy can cause these methods to produce sparse reconstructions, which can lead to inaccurate registrations. %Also, SfM or SLAM methods will probably fail in cases where the camera remains still and deformation occurs constantly, which is often the case in some applications like surgical force prediction. While our method, which can estimate dense reconstruction from a single frame, is still applicable.

Mahmoud et al.~\cite{Nader Mahmoud: DenseSLAM} propose a quasi-dense SLAM method for minimally invasive surgery that is able to produce dense reconstructions. However, it requires careful manual parameter tuning. Further, the accuracy of the reconstruction is lower than that required for sinus surgery, where low prediction errors are critical due to the proximity of critical structures such as the brain, eyes, carotid arteries, and optic nerves. Shape from Shading (SfS) based methods explicitly~\cite{Tatematsu:Shape,Ciuti:SfS} or implicitly~\cite{Austin:Learning} model the relationship between appearance and depth. These methods generally require \textit{a priori} modeling of the lighting conditions and surface reflectance properties. Since the true lighting and reflectance conditions are hard to model, SfS-based methods rely on simplified models that can result in noisy and inaccurate reconstructions, e.\,g., in the presence of specular reflections. 

Convolutional neural networks (CNNs) have shown promising results in high-complexity problems including general scene depth estimation~\cite{Iro:Deeper} which benefits from local and global context information and multi-level representations. However, using CNNs directly in endoscopic videos poses several challenges. First, dense ground truth depth maps are hard to obtain inhibiting the use of fully supervised methods. Hardware solutions, such as depth or stereo cameras, often fail to acquire dense and accurate depth maps from endoscopic scenes because of the non-Lambertian reflectance properties of tissues and paucity of features. Software solutions, such as those discussed above, do not produce reconstructions with the density or accuracy required for our application. More recent CNN-based methods~\cite{Marco: Deep monocular 3D reconstruction in bronchoscopy} use untextured endoscopy video simulations from CT to train a fully supervised depth estimation network and rely on another trained transcoder network to convert RGB video frames to texture independent frames required for depth prediction. This procedure requires per endoscope photometric calibration and complex registration which may only work well in narrow tube-like structures. It is unclear whether this method will work on in-vivo images since it is only validated on two lung nodule phantoms. Second, endoscopic images do not provide the photo-constancy that is required by unsupervised methods for depth estimation of general scenes~\cite{Tinghui:Unsupervised}. This is because the camera and light source move jointly and, therefore, the appearance of the same anatomy can vary substantially with different camera poses. In addition, texture-scarce regions make it hard to provide valuable information to guide the unsupervised network training even if the appearance was preserved across camera poses. 

In this work, we present a self-supervised approach to training deep learning models for dense depth map estimation from monocular endoscopic video data. Our method is designed to leverage improvements in SfM- or SLAM-based methods since our network training exploits reconstructions produced by these methods for self-supervision. Our method also uses the estimated relative camera poses to ensure depth map consistency in the training phase. While this approach requires the intrinsic parameters of the corresponding endoscope, it does not require any manual annotation, scaling, registration, or corresponding CT data.

\section{Methods}
We introduce a method for dense depth estimation in unlabeled data by leveraging established multi-view stereo reconstruction methods. Although SfM-based methods are only able to produce sparse reconstructions from endoscopic video data, these reconstructions and relative camera poses have been shown to be reliable~\cite{Simon:SfM}. Therefore, we use these reconstructions and camera poses to supervise the training of our network using novel loss functions. Doing so enables us to produce reliable dense depth maps from single endoscopic video frames.

\subsection{Training Data}
Our training data consists of pairs of RGB endoscopic images, 3D reconstructions and coordinate transformations between the image pairs from SfM, and the rectified intrinsic parameters of the endoscope. The training data generation is completely autonomous given the endoscopic and calibration videos and could, in principle, be computed on-the-fly with SLAM-based methods.

For each frame, we compute a sparse depth map to store the 3D reconstructions. By applying perspective geometry, 3D points can be projected onto image planes. Since SfM- or SLAM-based methods do not consider all frames when triangulating one particular 3D point, we only project the 3D points onto associated image planes. $b_{i,j}=1$ indicates frame $j$ is used to triangulate the 3D point $i$ and $b_{i,j}=0$ indicates otherwise. $\left(u_i^{j}, v_i^{j}\right)$ are projected 2D coordinates of the 3D point $i$ in frame $j$. The sparse depth map $Y_j^*$ of frame $j$ is 
\begin{equation}
    Y_j^*\left[ \left[ v_i^{j} \right], \left[ u_i^{j} \right] \right] = 
    \begin{cases} 
    z_i^{j} \quad \mbox{if} \, b_{i,j}=1\\
    0 \quad \mbox{if} \, b_{i,j}=0
    \end{cases} \quad \mbox{, where}
\end{equation}
$z_i^{j}$ is the depth of 3D point $i$ in frame $j$. Since the reconstruction is sparse, large regions in $Y_j^*$ will not have valid depth values.

We also compute sparse soft masks to ensure that our network can be trained with these sparse depth maps and mitigate the effect of outliers in the 3D reconstructions. This is achieved by assigning confidence values to valid regions in the image while masking out invalid regions. Valid regions are 2D locations on image planes where 3D points project onto, while the remaining image comprises invalid regions. The sparse soft mask, $W_j$, of frame $j$ is defined as
\begin{equation}
    W_j\left[ \left[ v_i^{j} \right], \left[ u_i^{j} \right] \right] = \\
    \begin{cases}
            c_i \quad \mbox{if} \, b_{i,j}=1\\
            0 \quad \mbox{if} \, b_{i,j}=0
    \end{cases} \quad \mbox{, where}
\end{equation}
$c_i$ is a weight related to the number of frames used to reconstruct 3D point $i$ and the accumulated parallax of the projected 2D locations of this point in these frames. Intuitively, $c_i$ is proportional to the number of frames used for triangulation and the accumulated parallax. Greater magnitudes of $c_i$ reflect greater confidence.

\subsection{Network Architecture}

\begin{figure}[t]
	\centering
	\includegraphics[scale=0.35]{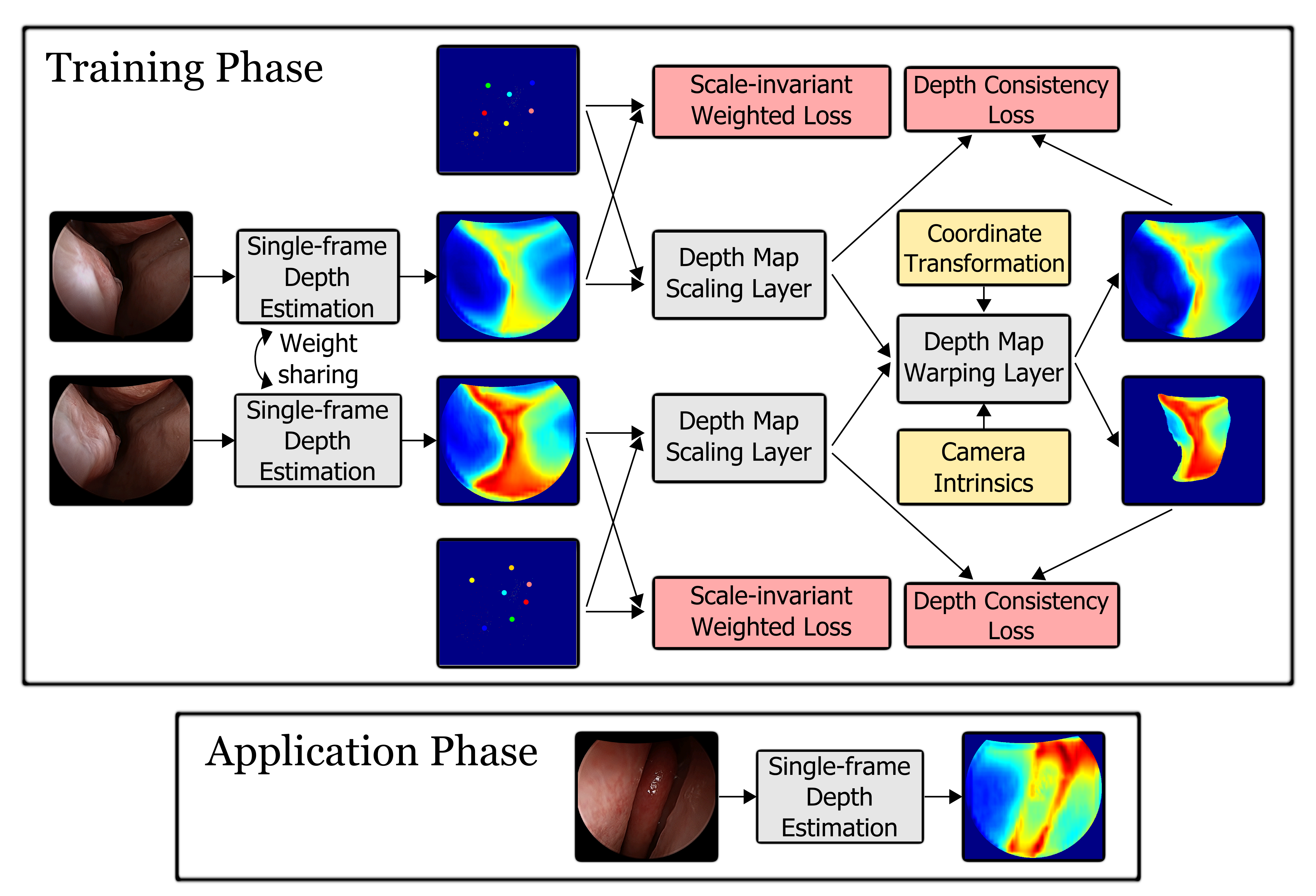} 
	\caption{Network architecture: our training network (top) is a self-supervised two-branch Siamese network that uses sparse 3D points and relative camera poses from SfM to estimate dense depth maps from pairs of images and enforce depth consistency, respectively. The soft sparse mask and sparse depth map are represented as a single blue square with dots. During the application phase (bottom), we use the trained weights of the single-frame depth estimation architecture (Fig.~\ref{fig:single-frame depth estimation}) to predict a dense depth map that is accurate up to a global scale.
	}
	\label{fig:overall_architecture}
\end{figure}

Our overall network architecture (Fig.~\ref{fig:overall_architecture}) is a two-branch Siamese network~\cite{Sumit:Siamese} with high modularity. For instance, our single-frame depth estimation architecture can be substituted with any architecture that produces a dense depth map. We introduce two custom layers in this network architecture.

The \textit{Depth Map Scaling Layer} scales the predicted dense depth map from the single-frame depth estimation architecture to remain consistent with the scale of the coordinate transformation. It uses the corresponding sparse depth map as the anchor point for scale computation.

The \textit{Depth Map Warping Layer} warps a scaled dense depth map to the coordinate frame of the other input to the Siamese network using the relative camera pose between the two frames. We implement this layer in a differentiable manner so that the training loss can be backpropagated. These two layers work together to generate data that is used to enforce depth consistency, described in the following section.  

\subsection{Loss Functions}

In the training phase, we use two loss functions that leverage the sparse depth annotations and relative camera poses between frames produced by SfM.

The first loss function, \textit{Scale-invariant Weighted Loss}, allows the network to train with sparse depth annotations because it uses sparse soft masks as weights to ignore regions in the training data where no depth values are available. Given a sparse depth map, $Y^*$, a predicted dense depth map, $Y$, and a sparse soft mask, $W$, the Scale-invariant Weighted Loss is defined as
\begin{equation}
   L_{sparse}\left(Y, Y^*, W\right) = \dfrac{1}{\sum_i w_i} \sum_i w_i d_i^2 - \dfrac{1}{\left(\sum_i w_i\right)^2} \left(\sum_i w_i d_i\right)^2 \quad \mbox{, where}
\end{equation}
$w_i$ is the value of the sparse soft mask at pixel location $i$ and $d_i = \log y_i - \log y_i^*$ is the difference between the predicted and ground truth depth at location~$i$~\cite{David:DepthMap}. The scale-invariance of this loss function
is advantageous given the inherent scale ambiguity of single-frame depth estimation.
It makes the network potentially generalizable to different patients, endoscopes, and anatomy because the network simply needs to estimate correct depth ratios without having to estimate the correct global scale. The global scale can vary considerably across different scenarios and is almost impossible for the network to estimate solely from endoscopic frames with no additional \textit{a priori} information as input. Finally, it makes the automatic training data generation in our method feasible. If the depth estimation network is set up to predict global scale, the results from SfM- or SLAM-based methods must resolve scale ambiguity first. This requires additional steps, e.\,g. registration to preoperative CT data, to recover the correct global scale. However, registration usually requires manual initialization and, therefore, user interaction. Alternatively, external tracking devices can record data that reflects global scale information but are often not accurate and can change the clinical workflow. With the Scale-invariant Weighted Loss, the automatically generated 3D reconstructions and camera poses are directly usable for network training. This allows our method to use all existing endoscopic videos as training data in a fully automatic manner as long as the intrinsic parameters of the corresponding endoscopes are known.

The second loss function, \textit{Depth Consistency Loss}, adds spatial constraints among frames in the training phase. By using the Scale-invariant Weighted Loss only, the network does not gain any information from regions where no sparse depth annotations are available and the training is prone to overfitting to the measurement noise or outliers from SfM- or SLAM-based methods. The Depth Consistency Loss helps gain more information and mitigate the overfitting issues. It requires inputs from the Depth Map Scaling Layer and the Depth Map Warping Layer. We denote the predicted depth map of frame $k$ as $Z_k$ and the warped depth map, warped from its original coordinate frame $j$ to the coordinate frame $k$, as $\check{Z}_{k,j}$. Pixels in $\check{Z}_{k,j}$ and $Z_k$ at location $i$ are denoted $\check{z}_i^{k,j}$ and $z_i^k$, respectively. The Depth Consistency Loss of frame $j$ w.\,r.\,t. $k$ is defined as
\begin{equation}
    L_{consist}\left(j, k\right) = \dfrac{1}{N} \sum_{i=1}^N \lvert\check{z}_i^{k,j} - z_i^k\rvert   \quad \mbox{, where}
\end{equation}
$N$ is the number of pixels in the region where both maps have valid depths.

The network overall loss is a weighted combination of the two loss functions defined above. Given the predicted dense depth map, $Y$, and sparse depth map, $Y^*$, the overall loss for network training with a single pair of training data from frame $j$ and $k$ is defined as
\begin{equation}
    \begin{aligned}
        L\left(j, k\right) = L_{sparse}& \left(Y_j, Y_j^*, W_j\right) + L_{sparse}\left(Y_k, Y_k^*, W_k\right) \\
        & + \omega \left(L_{consist}\left(j, k\right) + L_{consist}\left(k, j\right)\right) \quad \mbox{, where}
    \end{aligned}
\end{equation}
$\omega$ is used to control how much weight each type of loss function is assigned.

\section{Experimental Setup}

\begin{figure}[t]
	\centering
	\includegraphics[scale=0.3]{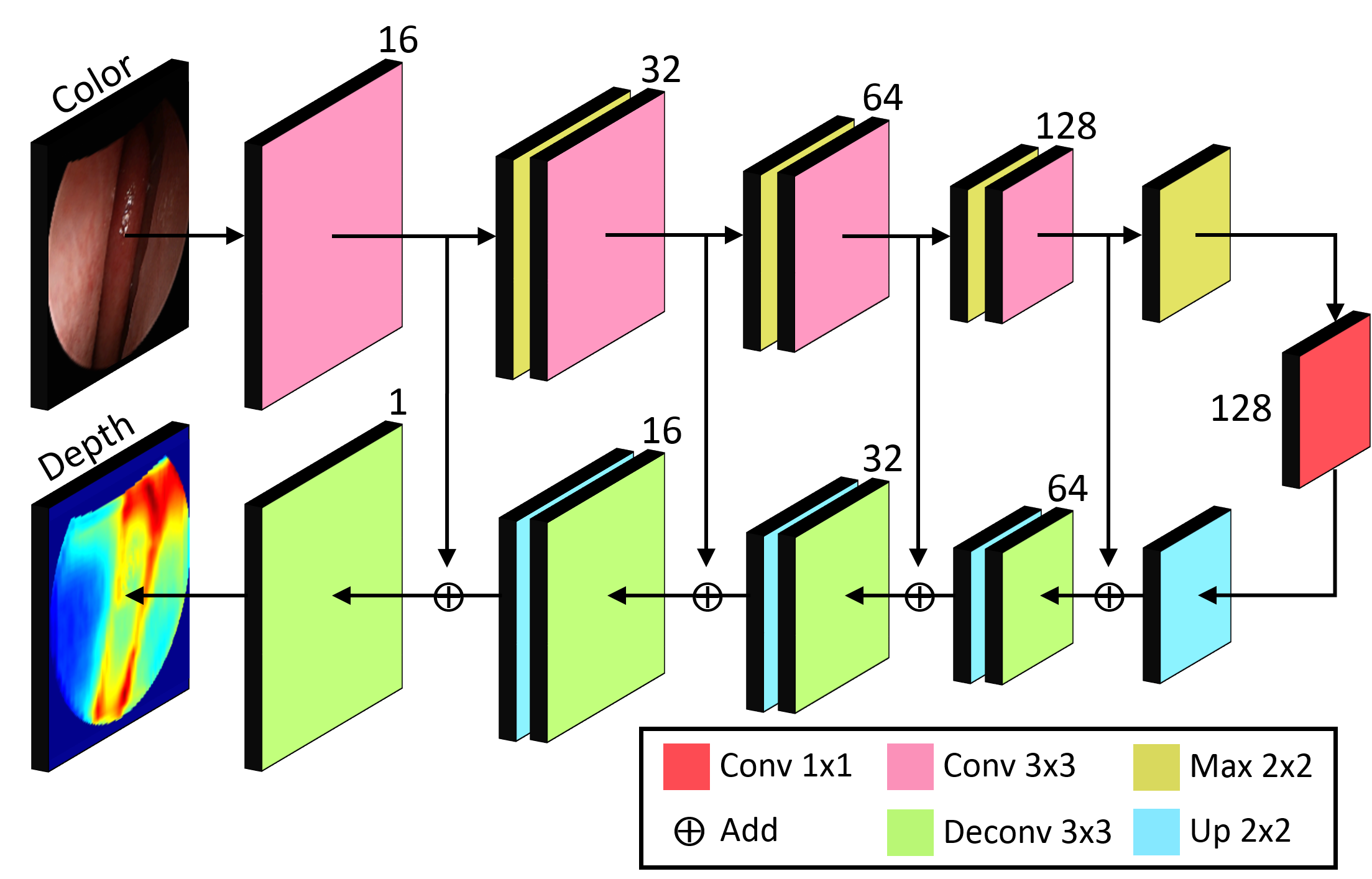}
	\caption{Single-frame depth estimation architecture: with the encoder-decoder architecture and symmetric connection skipping mechanism, the network is able to extract global information while preserving details.}
	\label{fig:single-frame depth estimation}
\end{figure}

Our network is trained using an NVIDIA TITAN X GPU with $12$GB memory. We use two sinus endoscopy videos acquired using the same endoscope. Videos were collected from anonymized and consenting patients under an IRB approved protocol. The training data consist of $22$ short video subsequences from Patient $1$. We use the methods explained above to generate a total of $5040$ original image pairs. The image resolution is $464\times512$, and we add random Gaussian noise to image data as an augmentation method. We use $95\%$ of these data for training and $5\%$ for validation. The learning rate and the weight, $\omega$, of the loss function are empirically set to $1.0\mathrm{e}^{-4}$ and $2.0\mathrm{e}^{-4}$, respectively. For evaluation, we use 6 different scenes from Patient $1$ and 3 scenes from Patient $2$, each containing 10 test images as input to the network in the application phase. These depth maps are converted to point clouds that were registered~\cite{Seth:GIMLOP} to surface models generated from corresponding patient CTs~\cite{Sinha17}. We use the residual error produced by the registration as our evaluation metric for the dense reconstructions. The single-frame depth estimation architecture we use is an encoder-decoder architecture with symmetric connection skipping (Fig.~\ref{fig:single-frame depth estimation})~\cite{Mao:Restoration}. %Note that the choice of depth estimation architecture is not the focus of this work and the proposed framework is modular, suggesting that the depth estimation network can easily be improved or replaced with other state-of-the-art architectures.
%% AS: This line does nothing but discredit your work - I don't think it is necessary

\section{Results and discussion}

\begin{figure}[t]
	\centering
	\includegraphics[scale=0.61]{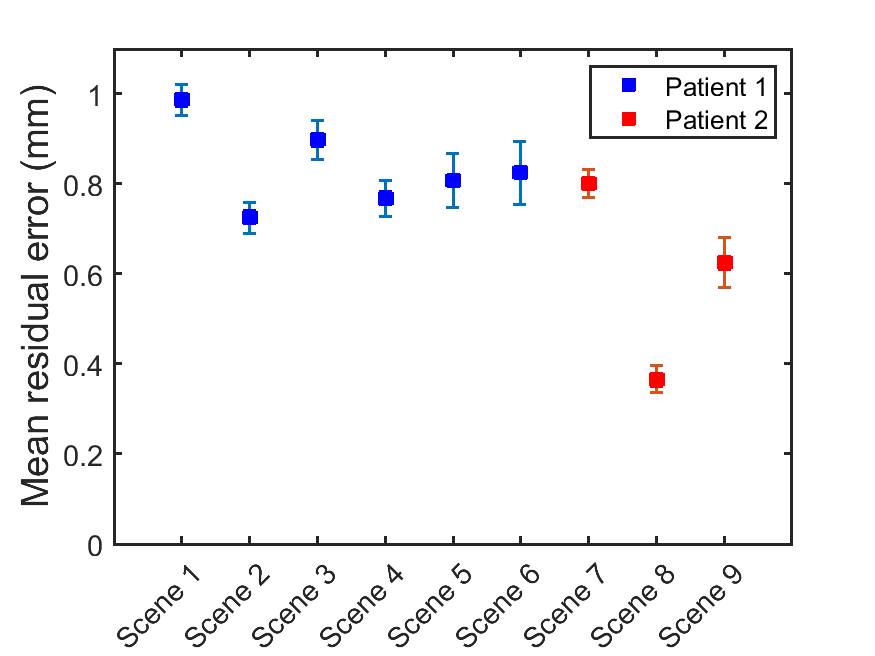}
	\caption{Mean residual errors for all testing scenes from Patients 1 and 2.}
	\label{fig:mean_residual_errors}
\end{figure}

The mean residual error produced by registrations over all reconstructions from Patient 1 is $0.84 \, (\pm \, 0.10)$\,mm and over all reconstructions from Patient $2$ is $0.63 \, (\pm \, 0.19$)\,mm. The mean residual error for Patient $1$ is larger than that for Patient $2$ due to the larger anatomical complexity in the testing scenes of Patient $1$. The residual errors for all $9$ testing scenes are shown in Fig.~\ref{fig:mean_residual_errors}. Since our method relies on results from SfM or other multi-view stereo reconstruction methods, improvements in these methods will be reflected immediately in our dense reconstructions. However, if these methods are not able to reconstruct any points from training videos or if the reconstructed points and estimated camera poses have large systematic errors, our method will also fail. 

We are able to detect and ignore frames where no reconstructions are estimated as well as individual outliers in reconstructions when the number of outliers is small relative to the number of inliers. However, there are cases where all reconstructed points and estimated camera poses are incorrect because of the extreme paucity of features in certain regions of the nasal cavity and sinuses. Currently, we rely on manual checking to ensure that 2D projections of SfM reconstructions are locked onto visual features in order to ignore erroneous reconstructions. However, in the future, we hope to develop an automatic method to detect these failures. Further, with training data from a single patient and evaluation on only two patients, it is unclear whether our method is able to generalize or is overfitting to this particular endoscope. Our current results also do not allow us to know whether or not fine-tuning the network in a patient-specific manner will improve the accuracy of reconstructions for that particular patient. In the future, we hope to acquire a larger dataset in order to investigate this further.

\begin{figure}[t]
	\centering
	\includegraphics[scale=0.15]{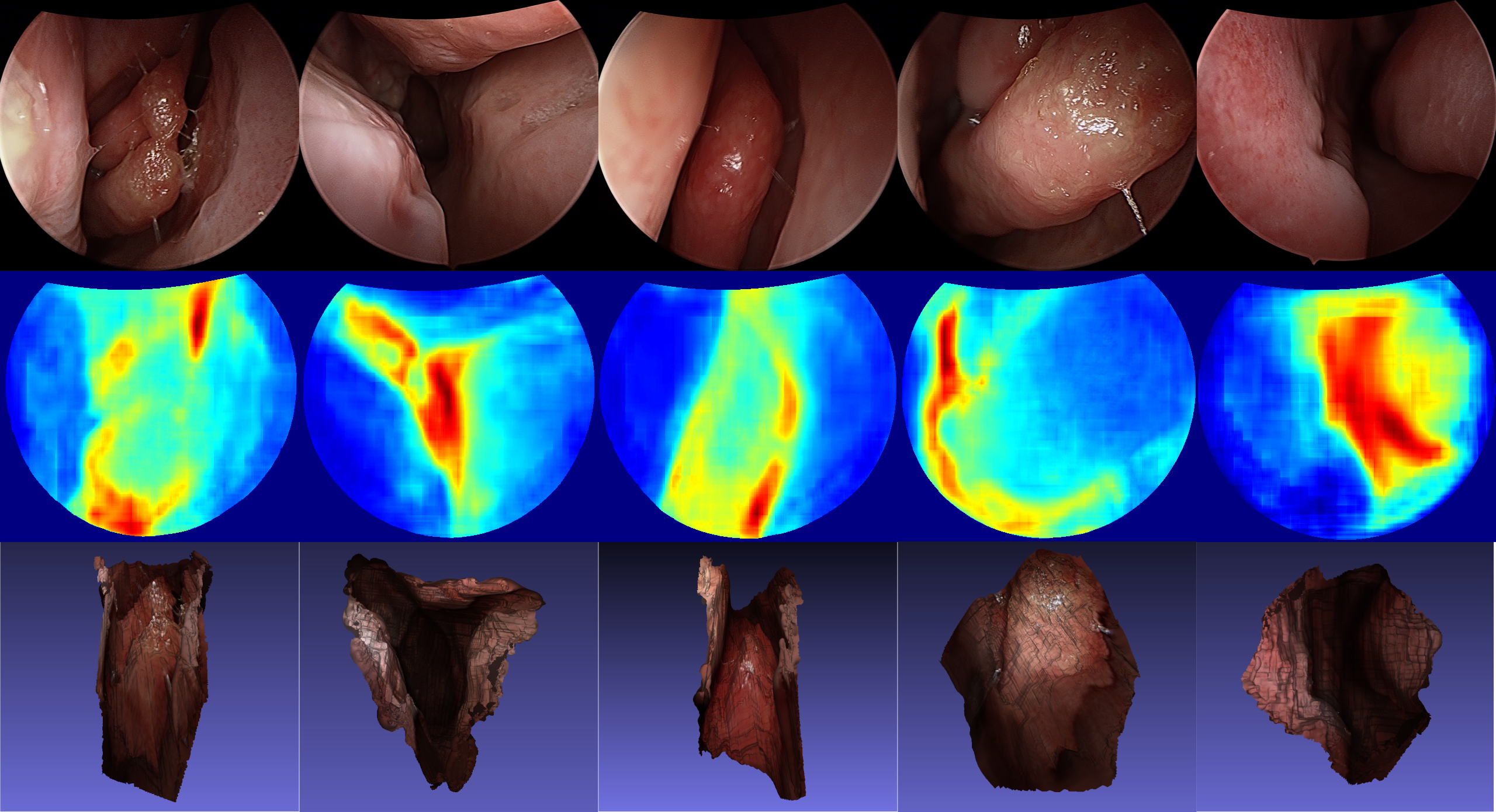}
	\caption{Examples of dense photometric reconstructions from Patients 1 and 2: 
	each column captures a different region in the nasal cavity and sinuses. The top row shows the color endoscopic images, the middle row shows the corresponding depth images where red maps to high values and blue to low values, and the bottom row shows the photo-realistic 3D reconstructions produced by our method.}
	\label{fig:results_showing}
\end{figure}

Samples from our current dense reconstruction results are shown in Fig.~\ref{fig:results_showing} for qualitative evaluation. There are several challenges in these examples where the traditional SfS methods are likely to fail. For example, shadows appear in the lower middle region of the second sample and the upper right region of the fourth sample. There are also specular reflections from mucus in the first, third and fourth samples. With the capability of extracting local and global context information, our network recognizes these patterns and produces accurate predictions despite their presence. Fig.~\ref{fig:comparison} also shows a comparison between a sparse reconstruction obtained using SfM and a dense reconstruction obtained using our method.

\section{Conclusion}
In this work, we present an approach for dense depth estimation in monocular endoscopy data that does not require manual annotations for training. Instead, we self-supervise training by computing sparse annotations and enforcing depth prediction consistency across multiple views using relative camera poses from multi-view stereo reconstruction methods like SfM or SLAM. Consequently, our method enables training of depth estimation networks using only endoscopic video, without the need for CT data, manual scaling, or labeling. We show that this approach can achieve submillimeter residual errors on sinus endoscopy data. Since our method can generate training data automatically and directly maps original endoscopic frames to dense depth maps with no \textit{a priori} modeling of anatomy or shading, more unlabeled data and improvements in SfM- or SLAM-based methods will directly benefit our approach and enable translation to different endoscopes, patients, and anatomy. This makes our method a critical intermediate step towards accurate endoscopic surgical navigation. In the future, we hope to evaluate our method on different endoscopes, patients, and anatomy and compare with other methods. Substituting the single-frame depth estimation architecture with a multi-frame architecture is also a potential future direction to explore.

\section*{Acknowledgement}
The work reported in this paper was funded in part by NIH R01-EB015530, in part by a research contract from Galen Robotics, and in part by Johns Hopkins University internal funds.

\end{document}